\documentclass[10pt,letterpaper]{article}

\usepackage{cogsci}
\usepackage{graphicx}

\cogscifinalcopy %%% Uncomment this line for the final submission

\usepackage[
  style=apa,
  natbib=true,
  annotation=false,
]{biblatex}
\usepackage{float} %%% Roger Levy added this and changed figure/table placement to [H] for conformity to Word template, though floating tables and figures to top is still generally recommended!

\title{LLMs and people both learn to form conventions -- just not with each other}

\author[1,4]{Cameron R. Jones}
\author[2]{Agnese Lombardi}
\author[3]{Kyle Mahowald}
\author[4]{Benjamin K. Bergen}
\affil[1]{Department of Psychology, Stony Brook University}
\affil[2]{Department of Philology, Literature, and Linguistics, University of Pisa}
\affil[3]{Department of Linguistics, University of Texas at Austin}
\affil[4]{Department of Cognitive Science, University of California San Diego}

\bibliography{manual}

\begin{document}

\maketitle

\begin{abstract}
Humans align to one another in conversation -- adopting shared conventions that ease communication. We test whether LLMs form the same kinds of conventions in a multimodal communication game. Both humans and LLMs display evidence of convention-formation (increasing the accuracy and consistency of their turns while decreasing their length) when communicating in same-type dyads (humans with humans, AI with AI). However, heterogenous human-AI pairs fail---suggesting differences in communicative tendencies. 
In Experiment 2, we ask whether LLMs can be induced to behave more like human conversants, by prompting them to produce superficially humanlike behavior.
While the length of their messages matches that of human pairs, accuracy and lexical overlap in human-LLM pairs continues to lag behind that of both human-human and AI-AI pairs.
These results suggest that conversational alignment requires more than just the ability to mimic previous interactions, but also shared interpretative biases toward the meanings that are conveyed.
%While humans and LLMs share …

\end{abstract}

\section{Introduction}

Humans align with one another in conversation \citep{garrod2004conversation}. They come to use vocabulary \citep{brennan1996conceptual}, phrases \citep{garrod1987saying}, and syntactic constructions \citep{branigan2000syntactic} that their partners have used more often.
This could be because people are cooperative communicators \citep{lewis2008convention} who follow Grice’s cooperative principle to be informative, concise, and on-topic \citep{grice1975logic}. Conversational alignment allows communicators to eliminate redundancy and develop a concise and effective set of referring expressions which are guaranteed to be familiar and salient to both parties \citep{wilkes1992coordinating}.
However, carrying out this kind of explicit cooperative alignment requires maintaining complex representations of partners' mental states and recursively tracking beliefs about what one's partner knows about oneself \citep{clark1996using}.

Other proposed accounts allow alignment to emerge through simpler mechanisms like frequency heuristics \citep{barr2004establishing}, priming mechanisms \citep{menenti2012toward, chartrand1999chameleon}, and first-order social rules \citep{hawkins2017convention}.
However, debate continues about the extent to which simpler statistical mechanisms could account for such rich interactive behavior \citep{fusaroli2016investigating, pickering2004toward}.

Large language models (LLMs)---neural networks that learn to predict words from linguistic context---are increasingly being deployed in multi-turn conversations with human partners. These models have proven effective at displaying humanlike behavior in short interactions \citep{jones2025large} and engaging millions of consumers \citep{chatterji2025people}.
However, it is unclear whether they develop same kinds of conversational alignment that people do with each other.

On one hand, we might expect that these systems would excel at adapting to their conversational partner in this way.
LLMs are especially effective at In-Context Learning (ICL): adapting to patterns that appear in their immediate context during inference without the opportunity to encode these patterns into their weights \citep{brown2020language, lampinen2025generalization}.
%\citep{mccoy} have argued that this ability accounts for many of LLMs' linguistic competencies...
Moreover, LLMs are explicitly trained to be cooperative communication partners via reinforcement learning from human feedback \citep{bai2022training}.
If people prefer turns that follow Gricean maxims, human feedback could bias models towards aligning to the user's expressions.

%Together, LLMs' in context-learning abilities and the training signal they receive from RLHF could provide both the capacity and tendency toward convention formation with human partner.
% IBM paper?
% Papers in federico chat?

\begin{figure*}[ht]
\begin{center}
\includegraphics[width=0.99\linewidth]{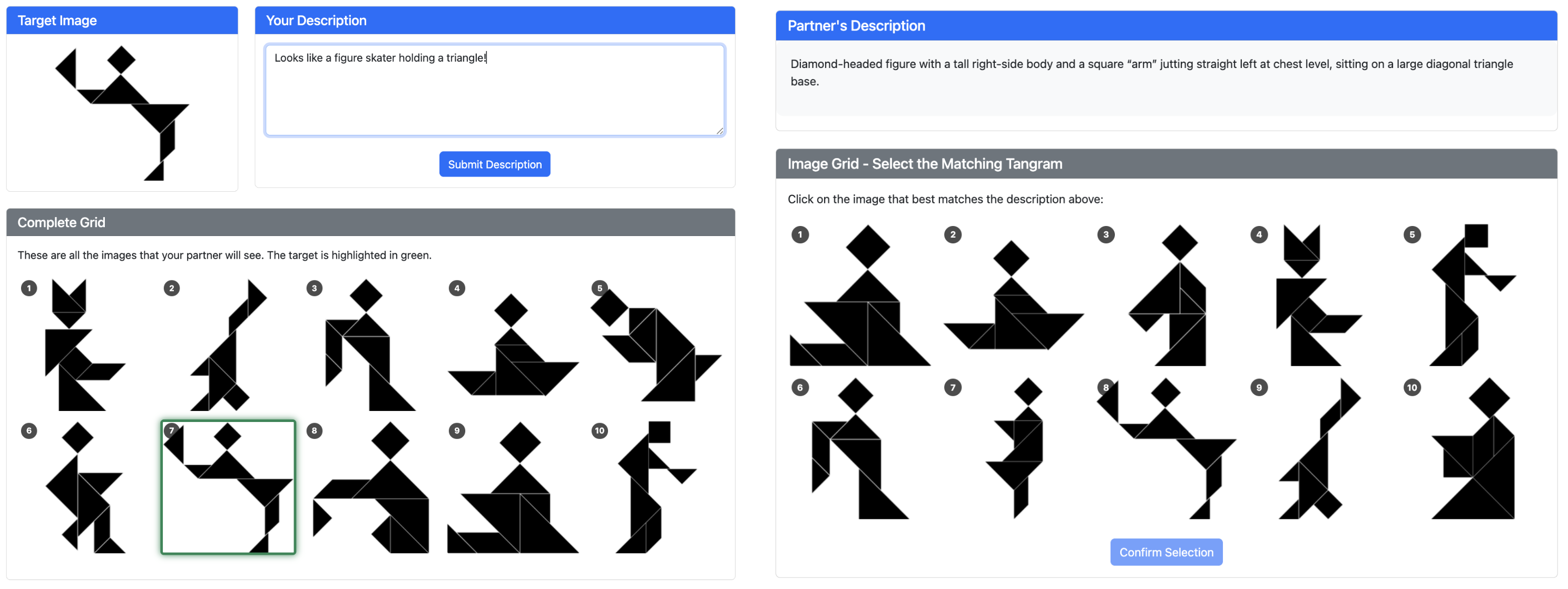}
\end{center}
\caption{The interface from the perspective of the human director (left) and the matcher (right). As directors, participants see a grid of figures and are instructed to describe the designated figure (in the green box) such that their partner, the matcher, will select it. The matcher is instructed to select the figure described by the director from the presented grid.} 
\label{fig:interface}
\end{figure*}

On the other hand, there are reasons to be sceptical that LLMs adopt humanlike conventions.
First, these models lack humans' resource-sensitivity.
While speaking and listening are inherently costly for people, LLMs have no equivalent natural pressure to be concise \citep{clark1991grounding, zipf1949human}.
Second, many authors have argued that LLMs lack the kind of communicative intent \citep{bender2020climbing, shanahan2024talking} which might be necessary for developing common ground \citep{lewis2008convention}.
Lastly, the development of common ground might require a shared interpretative stance on the world---compatible ways of perceiving and categorizing referents that allow partners to build on each other's descriptions \citep{brennan1996conceptual}.
Humans and LLMs might lack this due to differences in how they process and represent information \citep{mahowald2024dissociating}.

% LLMs might be capable but humans might not form with them...

To the extent that LLMs develop humanlike conventions, it could suggest that purely statistical learning (from distributional language statistics and reinforcement feedback) is sufficient to facilitate conversational alignment.
In contrast, if LLMs are incapable of this kind of alignment, it could suggest that more fundamental differences between human and LLM communicators is crucial for establishing common ground.

Here we investigate whether people and LLMs form shared conventions when they communicate in a version of the tangrams reference game \citep{wilkes1992coordinating}. In the game, partners communicate about a grid of ambiguous figures made from sets of simple geometrical shapes called tangrams. 
Human participants' descriptions tend to become shorter, more effective, and more stable over multiple rounds, indicating that they are forming conventions \citep{wilkes1992coordinating, hawkins2017convention}.
In Experiment 1, we compare Human-Human, Human-AI, and AI-AI pairs. After finding differences across these kinds of pairs, in Experiment 2 we modify the prompt for the AI partner to determine whether LLMs can be induced to behave more like human conversants through prompting.

Our work follows two recent studies which ask whether LLMs can form conventions. \citet{vaduguru2025success} found that pre-trained Vision Language Models failed to develop conventions in dyadic interactions without explicit training. However, when models were optimized for both communicative success and message cost, conventions emerged with human listeners. Critically, their evaluation was unidirectional—the model always spoke while humans listened.
\citet{zeng2026lvlms} used a similar paradigm to ours---a referential communication game about images of baskets---and compared human-AI dyads to pairs of humans and AIs.
Their task differed from ours in that one partner was the director throughout the experiment rather than swapping roles each turn.
They found that both AI-AI and human-AI pairs failed to develop conventions, with AI-AI accuracy dropping across rounds.
Our work extends this literature by testing bidirectional coordination with general-purpose chat models, and by asking whether prompting models to behave more humanlike can close the gap with human-human pairs.

\begin{figure*}[ht]
\begin{center}
\includegraphics[width=0.99\linewidth]{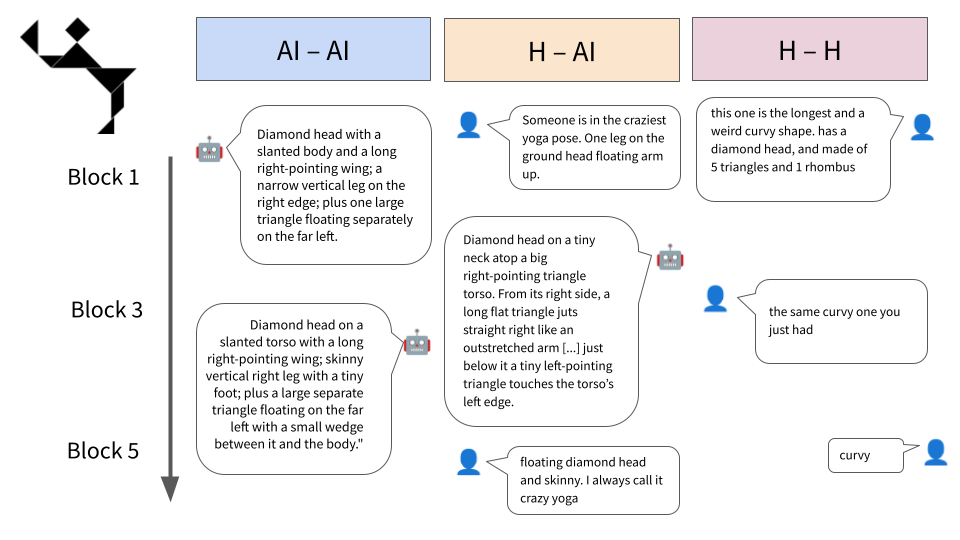}
\end{center}
\caption{Examples of turns across Experiment 1 from each partnership type. AI-AI pairs (left) produce lengthy detailed geometric descriptions which do not reduce over turns. H-H pairs (right) quickly form short and analogic conventions. H-AI pairs (middle) often showed a contrast, with human partners proposing conventions that were not adopted.} 
\label{fig:examples}
\end{figure*}

\section{Experiment 1}

In Experiment 1, we compared three different types of partnership: Human-Human, Human-AI, and AI-AI. Each pair communicated about the same grid of 10 tangrams over 50 turns. We used this setup to address four pre-registered research questions \citep{jones2025e1prereg}.

First, we asked whether each type of partnership would develop conventions over the course of the experiment as measured by a) their accuracy, b) their length, and c) the degree of lexical overlap between descriptions of the same image across rounds. On the basis of prior findings with human-human versions of this task \citep{hawkins2017convention, wilkes1992coordinating}, we predicted that each of these metrics would improve across the experiment for human-human pairs. If statistical learning from next-token prediction and the attendant in-context learning observed in LLMs is sufficient to generate humanlike conversational alignment, then we would expect that accuracy, length, and consistency across rounds should also improve in AI-AI and human-AI pairs. 

Second, we asked whether some types of partnership perform better than others.
To the extent that LLMs lack humanlike resource-consciousness, communicative intent, or representational biases, we should expect that H-AI pairs will  fall behind H-H pairs on all three metrics of convention formation.
Different theories make contrasting predictions about AI-AI pairs.
On one hand, if forming conventions requires recursively tracking interlocutors' mental states (at which humans outperform LLMs;  \cite{jones2024does}), then we might expect H-AI pairs (with at least one human) to best AI-AI pairs (with none).
Alternatively, if convention-formation rests on sharing conceptual and representational biases about how to carve up meaning spaces, then we might expect homogenous AI-AI pairs to outperform heterogenous H-AI ones.

Third, we asked whether human partners would adapt better to human partners: specifically whether they would produce shorter or more conventional messages to human partners than they did to AI partners. This would suggest that human behavior at the task is flexible and sensitive to partner behavior.

Lastly, we asked about participants' self-reported perception of human and AI partners in an exit survey. We predicted that humans would rate other humans as more helpful, adaptable, and more likely to be human than AI partners.

\begin{figure*}[ht]
\begin{center}
\includegraphics[width=0.99\linewidth]{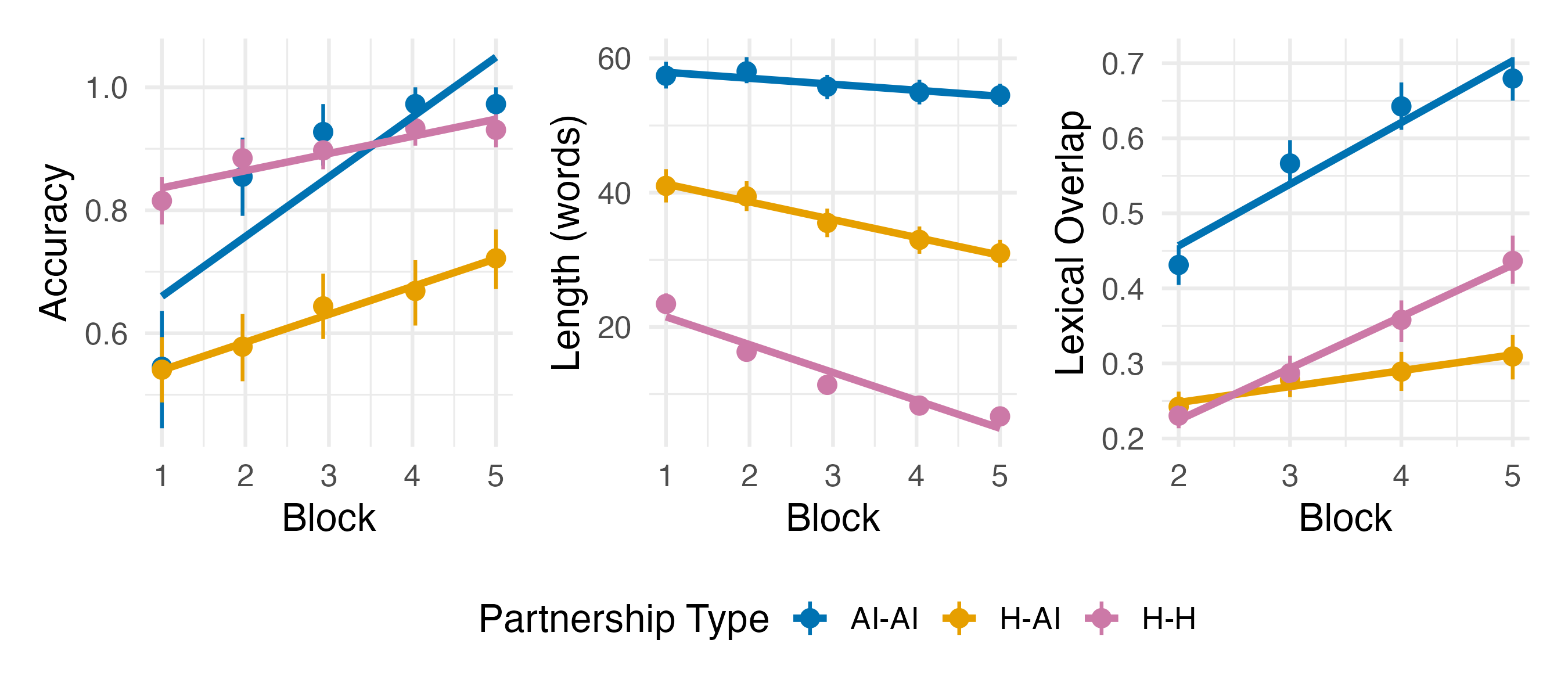}
\end{center}
\caption{ \textbf{Left:} Accuracy increased for all partnership types in Experiment 1. AI partnerships all started below H-H pairs, but while AI-AI pairs quickly caught up, H-AI pairs did not. \textbf{Center:} H-H messages started shorter than other pair types and also decreased fastest.
\textbf{Right:} Lexical overlap between descriptions of the same object from one block to the next increased for all partnership types. H-H pairs showed more overlap than H-AI pairs but less than AI-AI pairs.
}
\label{fig:e1_3_panel}
\end{figure*}

\subsection{Method}

\subsubsection{Materials}

We used a sample of 18 humanoid tangram figures from \citet{hawkins2017convention}. We used GPT-5 as the AI model in these experiments. We accessed the model through the OpenAI API between 2025-10-13 and 2025-10-24 with reasoning effort set to ``low''. The model prompt was identical to the instructions that human participants saw, with additional information about how responses should be formatted.
% Info about image encoding?

\subsubsection{Participants}

We recruited 168 participants through our institution's undergraduate subject pool. 59 participants failed to complete the experiment (for example because either they or their partner disconnected during the experiment).
No participants met our exclusion criteria of having average message lengths of less than 10 characters or average response times of less than 5s.
Overall we retained data from 109 participants.

\subsubsection{Procedure}

Participants completed the study online. After providing informed consent, they were randomly assigned to a condition (having either a human or AI partner) and entered into a waiting room. Participants in the human-human condition were matched up with the first available partner.
Participants in the human-AI condition waited for a random delay to simulate the matchmaking process.
Participants were not informed whether they were interacting with a human or an AI system and the instructions used the neutral term 'partner'.

Participants completed 50 rounds with the same partner, alternating director and matcher roles (see Figure \ref{fig:interface}). Directors described a cued image and matchers attempted to selected the described image from a grid. Both received feedback about which image was cued and selected after each round. Rounds were organized into 5 blocks of 10 images, so each image appeared once per block.

After all 50 rounds, participants completed an exit survey, rating their partner using a 7 point likert scale on i) how helpful they were, ii) how much they took the participants' perspective, iii) how much they used shared history, and iv) how much they adapted to the participant. Finally, participants were asked whether they thought their partner was human or an AI and gave a confidence score (0-100) and their reasoning.

% AI-AI games
AI-AI games were conducted using the exact same setup as for AI partners in human games, except that the AI system was matched with another copy of the AI system.

\subsection{Results}

For each trial, we measured i) whether the matcher accurately selected the intended image, ii) the length of the description in words, iii) and the jaccard similarity between the words in the description of an image and the pair's description of the same image in the previous block (lexical overlap).

% Add pvals / stats?

As predicted, each of the three partnership types significantly increased in accuracy and lexical overlap, while decreasing in length across blocks
(see Figure \ref{fig:e1_3_panel}).
However, these trends were qualitatively very different. H-H pairs accuracy started high (81\%) and increased steadily to 93\% by block 5. H-AI accuracy started much lower (54\%) and increased at a similarly steady rate to H-H pairs. 
AI-AI accuracy also started low (56\%) but shot up to 86\% by Block 2 and had reached ceiling (99\%) by Block 4.
While the length of H-H pairs started at 20 words and dropped to 6, AI-AI pairs' messages dropped only from 58 to 56, and H-AI pairs' messages were somewhere in between.
AI-AI pairs started with the highest lexical overlap and maintained this lead across all blocks. While H-AI and H-H pairs started with similar overlap ($\sim 0.23$), H-H pairs steadily increased to 0.43 by Block 5 while H-AI pairs lagged at 0.3.

H-H pairs outperformed H-AI pairs on all three metrics, as predicted (all $p < 0.009$). However, the AI-AI to H-AI comparison produced mixed results. While H-AI messages were shorter than AI-AI pairs' ($t = -6.36$, $p < 0.001$), AI-AI pairs were both more accurate and had higher lexical overlap with previous turns (both $p < 0.001$).

As predicted, human partners rated other humans significantly higher on each of the exit questions (all $p < 0.001$). While 83\% of participants believed that GPT-5 partners were AI, just 12\% of participants thought this of human partners.

As an exploratory analysis, we used GPT-5-mini to tag descriptions as either analogical (using analogies such as 'the ice skater' to refer to images), geometric (e.g. describing the image in terms of shapes), or a mixture.
68\% of H-H descriptions were tagged as analogical, versus 12\% as geometric and a further 13\% as mixed. Just 5\% of AI-AI pairs' messages were classed as analogical, with 38\% being classed as geometric and 57\% as mixed.
H-AI messages were 24\% analogical, 23\% geometric and 52\% mixed.

\begin{figure*}[ht]
\begin{center}
\includegraphics[width=0.99\linewidth]{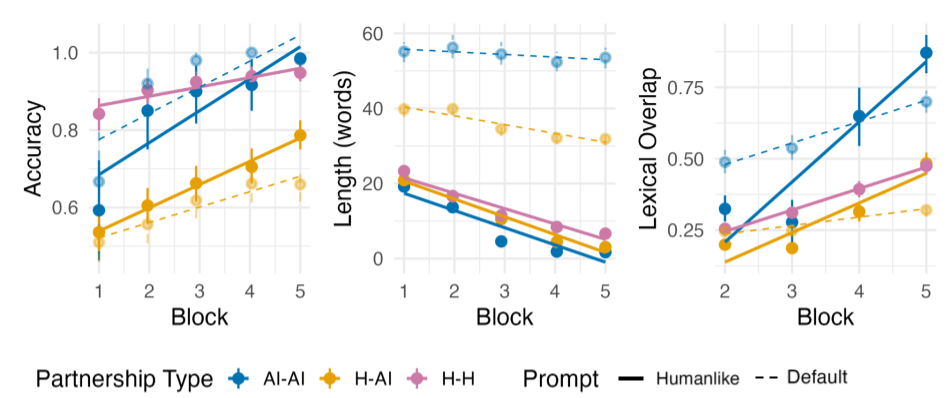}
\end{center}
\caption{Experiment 2: A new ``Humanlike'' prompt caused both H-AI and AI-AI pairs to produce a similar trajectory in description length to H-H pairs (center).
While AI-AI pairs showed all of the hallmarks of convention-formation, H-AI pairs continued to show significantly lower accuracy (left) and lexical overlap (right) than H-H pairs. } 
\label{fig:e2_3_panel}
\end{figure*}

In order to better understand the mechanism behind convention formation, we found the mean Jaccard similarity between descriptions of the same tangram in subsequent rounds, conditioned on whether or not the last description had been successful.
Curiously, we saw significant increases in similarity for successful vs unsuccessful previous turns for all director/matcher combinations (all $p < 0.001$) except for AI directors/human matchers ($p = 0.307$).

% Exploratory analyses
% - Within - between

\subsection{Discussion}

We set out to test whether LLM-based AI systems could establish common ground conventions in the same way that humans do. The results firmly replicated the finding that human communicators form these conventions with one another. H-H pairs showed the hallmark increases in accuracy and lexical overlap, as descriptions shrank to just a few words (Figure \ref{fig:examples}).

In contrast, H-AI pairs failed to match human performance, in line with recent work on similar tasks \citep{zeng2026lvlms}. While these partnerships showed small significant increases on each of the metrics, their performance remained qualitatively poor (sending 30 word messages with only 70\% success by Block 5).
What accounts for this failure?
We found that human speakers send longer messages to AI models than they do to other humans, which might not create the requisite downward pressure on length to facilitate conventionalization.
In our exploratory analyses, we found that AI systems' descriptions were qualitatively different from humans': predominantly geometric rather than analogical.
Moreover, we found that AI directors were not sensitive to the success of previous turns when producing descriptions of an image specifically when interacting with a human partner, which could account for the uniquely poor performance of H-AI pairs.

The most surprising performance came from AI-AI pairs. In contrast to \citet{zeng2026lvlms}, we found these pairs displayed evidence of aligning to one another, but in a very different way from humans. They reached ceiling performance on accuracy within three blocks, with twice the proportion of lexical overlap as H-H pairs by Block 5. 
Their low accuracy in Block 1 suggests that AI-AI pairs are constructing genuine context-driven conventions, rather than context-free similarity between partners.
On the other hand, however, the length of their messages barely dropped (from 58 to 56 words).

These results suggest that AI systems are capable of developing stable, shared forms of reference, but that they are not resource-sensitive in the same way that humans are. These results make sense with respect to models' architecture and training regime. 
The transformer architecture \citep{vaswani2017attention} allows models to attend to any previous token in the input, potentially allowing error-free recall of previous descriptions \citep{olsson2022context}. However, in contrast to humans who bear metabolic and opportunity costs for all of their actions, models have no intrinsic incentives to be concise. 

The fact that both AI-AI pairs and H-H pairs are able to achieve high accuracy rates raises the question of why H-AI pairs perform so much more poorly.
This could be due to inherent differences in how human and AI participants interpret images, or from more superficial features of their descriptions that prevent development of high lexical overlap.
We address this question in Experiment 2.

\section{Experiment 2}

% Prompting has been found to be effective ... TT?
To test whether the differences between Human and AI players could be attributed to stylistic differences, we designed a prompt which explicitly instructed the system to behave more like humans. We used AI-AI game simulations to iteratively adjust the prompt until it produced descriptions that were qualitatively similar to human ones and showed a similar trajectory in terms of length.
For example, descriptions for the image in Figure \ref{fig:examples} might start as ``leaping dancer — diamond head; separate big triangle top‑left (raised arm); long right‑pointing arm; skinny right leg of two small triangles with a tiny foot'' in block 1 and end as ``leaping dancer'' by block 5.

We replicated Experiment 1, comparing this 'Humanlike' prompt to the 'Default' prompt in the H-AI condition. We asked 3 pre-registered research questions \citep{jones2025e2prereg}. First, does the the Humanlike prompt outperform the Default prompt (in both the convention metrics and exit survey)? This would confirm the theory that the difference between H-H and H-AI pairs is partly accounted for by superficial features which can be easily addressed through instructing the model to behave differently.

Second, we asked whether H-H pairs would continue to outperform Humanlike-prompted H-AI pairs. Any residual performance gap between these conditions would point to more fundamental differences between humans and LLMs, which cannot be closed by prompting models to be more concise or use analogical descriptions.

Lastly, we asked whether human participants would be able to tell whether the Humanlike-prompted AI partner was not human: specifically whether they judged it to be an AI more often than would be expected if they were randomly guessing.
% Add motivation

\subsection{Methods}

The experiment proceeded identically to Experiment 1 except that participants were assigned to one of 3 types of partner: Human, AI (Default) and AI (Humanlike). AI (Default) was identical to the setup used in Experiment 1. AI (Humanlike) was also a GPT-5 model with the same parameter settings, but the prompt contained additional instructions to use analogical descriptions, to keep descriptions very short, and to lower the length of descriptions across the game. 
% Make full text available?
142 participants completed the experiment, and we retained 39 sessions of each of the H-AI partners and 32 H-H sessions.

\subsection{Results}

In H-AI games, the Humanlike prompt led to shorter ($t = -17.7$, $p < 0.001$) and more accurate turns ($z = 1.99$, $p = 0.04$) compared to the Default prompt (see Figure \ref{fig:e2_3_panel}). Lexical overlap increased numerically but not significantly ($t = 1.54$; $p = 0.126$). The Humanlike prompt was judged significantly higher on all of the exit survey questions (all $p < 0.001$).

Compared to H-H pairs, Humanlike-prompted H-AI pairs produced messages that were not significantly longer (and in fact, numerically shorter; $t = 1.58$, $p = 0.118$). However, H-H messages showed significantly higher lexical overlap ($t = 2.98$,  $p = 0.004$) and accuracy ($z = 9.34$,  $p < 0.001$). Moreover, human partners outperformed the Humanlike prompted AI partners across all exit survey questions (all $p < 0.02$).

84\% of participants with human partners believed they were human, while just 12\% of participants thought the Default-prompted AI model was human.
Partners of the Humanlike-prompted AI believed it was human 59\% of the time.
This rate was significantly higher than the Default prompt ($z = 3.63$, $p < 0.0001$) and crucially no different than would be expected if participants were randomly guessing ($p = 0.90$).
However, it was significantly lower than the rate at which humans were judged to be human ($p = 0.008$).

Exploratory analyses showed that the Humanlike prompt produced a similar distribution of description types to H-H pairs (53\% analogic, 12\% geometric, 30\% mixed).
Moreover, the Humanlike prompt showed a significant increase in lexical overlap in cases where the previous description for that image had been successful ($p < 0.001$).

\subsection{Discussion}

Experiment 2 tested a prompt addressing some of the superficial differences between human and LLM behavior in Experiment 1.
When LLMs were prompted in this way, human-AI pairs produced descriptions that were a similar length to the ones human-human pairs sent. However, H-AI pairs still showed lower lexical overlap and accuracy across the experiment than H-H pairs.
The results rule out the account that H-AI pairs fail to form conventions because AI systems lack sensitivity to message length.

AI-AI games that used this prompt achieved very high lexical overlap and accuracy, while achieving humanlike conciseness.
While the AI-AI results were not pre-registered (but part of the design of the experiment that we developed iteratively), they do demonstrate that LLMs are capable of producing convention-forming behavior with other LLMs--just not with humans. This implies that there are deeper differences between humans and LLMs which specifically limit convention formation.

\section{General Discussion}

Across two experiments, we found that same-type pairs (H-H and AI-AI) successfully formed conventions while mixed H-AI pairs consistently failed—even when models were prompted to produce humanlike behavior. The results have several implications for theories of convention-formation.
First, they complicate the question of what is required to form conventions: LLMs appear to be able to form conventions with one another, just not with humans.
That they can develop stable conventions suggests that they possess the requisite capabilities to learn in context.
The more challenging question is whether this suggests that LLMs possess deeper capacities for recursive reasoning and mental modeling that some argue are required for convention-formation \citep{grice1975logic, lewis2008convention, wilkes1992coordinating}, or that these capacities are not necessary after all \citep{barr2004establishing, hawkins2017convention, menenti2012toward}.
Future work using mechanistic approaches may inform whether humans and LLMs form conventions in similar ways.

Secondly, the results imply that forming conventions requires more than just the capacity to refer back to and repeat a partner's expressions \citep{brennan1996conceptual}. 
While AI-AI accuracy was low in block 0 (suggesting that LLMs also failed to produce effective initial descriptions for other LLMs), AI partners were able to agree on consistent conventions over later blocks while H-AI pairs were not.
Perhaps this is because AI models shared the same interpretative biases from their architecture, training data, and prompt.
Forming conventions appears to require some additional kind of shared interpretative stance: partners need to start with enough common ground to build on.
At a high level, these results suggest that the ability to form conventions is a property of a dyad---rather than of an individual \citep{clark1996using}.

Lastly, our results imply that the social interactions which people are having with LLMs will lack the interactive alignment and development of common ground that human conversations develop.
Moreover, we found that people in Experiment 2 couldn't reliably tell when they were talking to a suitably prompted AI system.
This suggests that people may not be sensitive to models' systematic failings--continuing to believe that these systems are humans even when they fail in very unhumanlike ways.

\printbibliography

\end{document}